\documentclass[final]{cvpr}

\usepackage{times}
\usepackage{epsfig}
\usepackage{graphicx}
\usepackage{amsmath}
\usepackage{amssymb}

\usepackage[ruled,vlined]{algorithm2e}
\usepackage[dvipsnames]{xcolor}
\newcommand\scalemath[2]{\scalebox{#1}{\mbox{\ensuremath{\displaystyle #2}}}}

\usepackage{multicol}
\usepackage{color}
\usepackage{comment}
\usepackage{booktabs}

\usepackage[pagebackref=true,breaklinks=true,colorlinks,bookmarks=false]{hyperref}

\definecolor{ben}{rgb}{0.9,0.,0.5}

\begin{document}

\title{NeRF-Pose: A First-Reconstruct-Then-Regress
Approach for \\ Weakly-supervised 6D Object Pose Estimation}

\author{
Fu Li$^{1,2*}$ \and Shishir Reddy Vutukur$^{2,3}$\thanks{The first two authors contributed equally to this work}
 \and Hao Yu$^{2}$ \and Ivan Shugurov$^{2,3}$  \and Benjamin Busam$^{2}$ \and  Shaowu Yang$^{1}$ \and Slobodan Ilic$^{2,3}$ \and \quad \quad
$^{1}$ National University of Defense Technology \quad
\quad $^{2}$ Technical University of Munich \\ 
$^{3}$ Siemens AG, Munich. 
}

\maketitle

\begin{abstract}\noindent
Pose estimation of 3D objects in monocular images is 
a fundamental and long-standing problem in computer vision.
Existing deep learning approaches for 6D pose estimation typically rely on the availability of 3D object models and 6D pose annotations.
However, precise annotation of 6D poses in real data is intricate, time-consuming and not scalable, while synthetic data scales well but lacks realism.
To avoid these problems, we present a weakly-supervised reconstruction-based pipeline, named \textbf{NeRF-Pose}, which needs only 2D bounding boxes and relative camera poses during training.  
Following the first-reconstruct-then-regress idea, we first reconstruct the objects from multiple views in the form of an implicit neural representation. 
Then, we train a pose regression network to predict pixel-wise 2D-3D correspondences between images and the reconstructed model.
A NeRF-enabled PnP+RANSAC algorithm is used to estimate stable and accurate pose from the predicted correspondences.
Experiments on LineMod and LineMod-Occlusion show that the proposed method has state-of-the-art accuracy in comparison to the best 6D pose estimation methods in spite of being trained only with weak labels. 
We extend the Homebrewed DB dataset with real training images to support the weakly supervised task and achieve compelling results. The extended dataset and code will be released soon.
\end{abstract}
\section{Introduction}
\begin{figure}
    \centering
    \includegraphics[width=\linewidth]{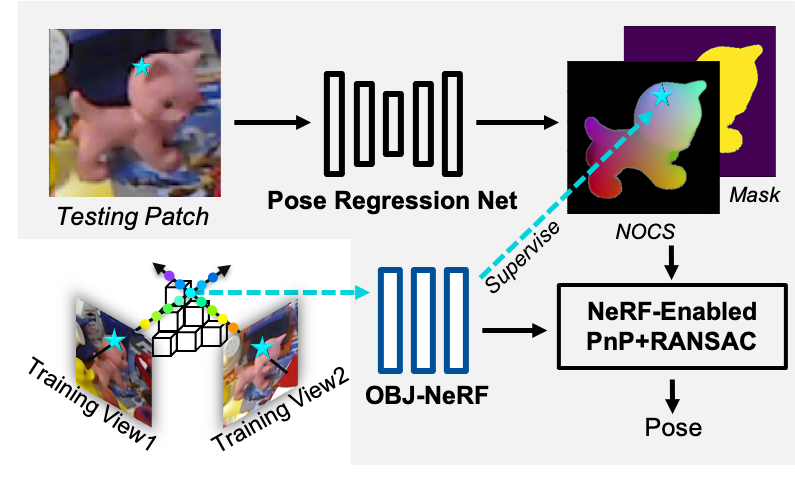}
        \caption{The inference of \textbf{NeRF-Pose}. 1) The pose regression network predicts a binary segmentation mask and 2D-3D correspondences represented as \textit{NOCS} maps. The \textit{g.t. NOCS} maps are generated from the OBJ-NeRF network, which encodes implicit 3D object representation recovered from the multi-view weak labels, \textit{i.e.,} 2D object segmentation masks and relative camera poses. 2) The proposed NeRF-enabled PnP+RANSAC is performed on the regressed results to predict the object pose.
        }
    \label{fig:teaser}
\end{figure}

Several computer vision tasks, such as 2D object detection and semantic segmentation, have experienced tremendous progress in recent years thanks to the development of deep learning. 
However, 2D object detection~\cite{girshick2015fast,he2017mask,law2018cornernet,liu2016ssd,redmon2016you,ren2015faster} alone is limited and insufficient for real-world applications such as Augmented Reality, Robotic Manipulation, Autonomous driving, etc., which often require the knowledge of the full 6 degrees of freedom (DoF) pose of the object in the scene. Therefore, the ability to recover the pose of the objects in 3D environments is essential for a better understanding of the 3D scene.

This fundamental 3D vision problem has been addressed by scholars for decades and is particularly difficult if the estimation is done only from a single RGB image.
Recent research tackles the ill-posed nature of this problem in a data-driven way where the pose is usually computed with respect to a 3D CAD model.
Most of the recent approaches~\cite{kehl2017ssd,li2018deepim,li2019cdpn,peng2019pvnet,rad2017bb8,sundermeyer2018implicit,tekin2018real,xiang2017posecnn,zakharov2019dpod,labbe2020cosypose,wang2021gdr,di2021so} require 6D pose labels as supervision signals. 
Moreover, most of the recent methods posed the pose estimation as a problem of correspondence estimation between a known 3D object model, such as a CAD model, and image pixels. However, it is hard and expensive to obtain accurate pose labels and fine-grained CAD models for all objects for real-world scenarios~\cite{kaskman2019homebreweddb}.
On the other hand, synthetically generated images have the advantage of a potentially unlimited amount of labeled data. However, methods trained only on synthetic data~\cite{kehl2017ssd,sundermeyer2020augmented,zakharov2019dpod,busam2020like,kaskman2020} have worse performance than their real counterparts due to the lack of realism. Moreover, precise textured CAD models are required to render synthetic data.
We argue that it is much easier to obtain 2D image labels, such as segmentation masks, and relative camera poses. As widely used in the recent methods~\cite{kehl2017ssd,li2018deepim,li2019cdpn,peng2019pvnet,rad2017bb8,sundermeyer2018implicit,tekin2018real,xiang2017posecnn,zakharov2019dpod,labbe2020cosypose,wang2021gdr,di2021so}, segmentation mask can be obtained either manually or automatically with the off-the-shelf object segmentation approaches~\cite{kirillov2023segment,he2017mask,chen2017deeplab}, few-shot segmentation\cite{wang2019panet,nguyen2019feature}, depth driven segmentation or background substaction\cite{garcia2020background},
while relative camera poses can be obtained with Structure from Motion~(SfM)~\cite{schonberger2016structure}, Simultaneous Localization And Mapping~(SLAM)~\cite{bailey2006simultaneous}, Inertial Visual Odometry~\cite{leutenegger2015keyframe}, or simply a marker board~\cite{kaskman2019homebreweddb}.
Different from fully-supervised methods trained on all labels, \textit{e.g.} CAD models, segmentation masks and 6D pose labels, the methods, that don't use textured CAD models and 6D pose annotations, use weaker labels and, so, can be considered as weakly-supervised.

Our approach is to first recover implicit 3D object representation from training images containing weak labels: 2D segmentation masks and relative camera poses. Next, we use this implicit representation to supervise the regression of dense correspondences between training images and previously recovered object's implicit 3D object representation. 
Therefore, we propose a \textit{first-reconstruct-then-regress} training pipeline, named \textbf{NeRF-Pose}, which builds atop of the success of Neural Radiance Fields~ (NeRF)~\cite{mildenhall2020nerf} and its successors~\cite{lin2021barf,yu2021plenoctrees,reiser2021kilonerf,yang2021objectnerf}. We first reconstruct the object as a NeRF-based network trained with weak labels. Then, we train a pose regression network to regress the dense image pixel~(2D)-object model~(3D) correspondences.

As depicted in Fig.~\ref{fig:teaser}, during inference, we first detect the objects in a 2D image using an off-the-shelf 2D object detection network and then predict segmentation masks and dense correspondences represented in terms of Normalized Object Coordinates~(\textit{NOCS})~\cite{Wang2019nocs}. With regressed correspondences and the NeRF object model, we propose a NeRF-enabled PnP+RANSAC method in order to compute the object pose in the end. Our key contributions can be summarized as follows:
\begin{itemize}
    \setlength{\itemsep}{0pt}
    \item A weakly-supervised object pose estimation approach, which is trained only with 2D annotations and relative camera poses, instead of relying on an explicit CAD model and accurate 6D pose labels.
    \item OBJ-NeRF neural network encoding an implicit 3D object representation obtained from weak labels: segmentation masks and relative camera poses.
    \item Pose regression network, which relies on the above object's implicit NeRF-based representation.
    \item NeRF-enabled PnP+RANSAC approach enabling highly accurate 6D pose computation.
    \item An extension of the HomebrewedDB dataset containing real video sequences with weak labels~(segmentation masks and relative camera poses).
\end{itemize}

We conduct experiments on LineMod~(LM)~\cite{hinterstoisser2012model} and LineMod Occlusion~(LMO)\cite{brachmann2014learning} datasets, and extend the HomebrewedDB~(HBD)\cite{kaskman2019homebreweddb} dataset. 
Though training in weakly supervised settings, we have about 15\%~(LM) and 20\%~(LMO) improvement on ADD(-S) metric compared to the methods trained without CAD models but with accurate pose labels. We also achieve comparable results on LM, LMO, and HBD datasets in comparison to fully supervised methods. The experiments show that our weakly-supervised NeRF-Pose approach achieves accurate and robust object pose estimation. 
\section{Related Work}\label{sec:related}

\begin{figure*}[t]
    \centering
    \includegraphics[width=\linewidth]{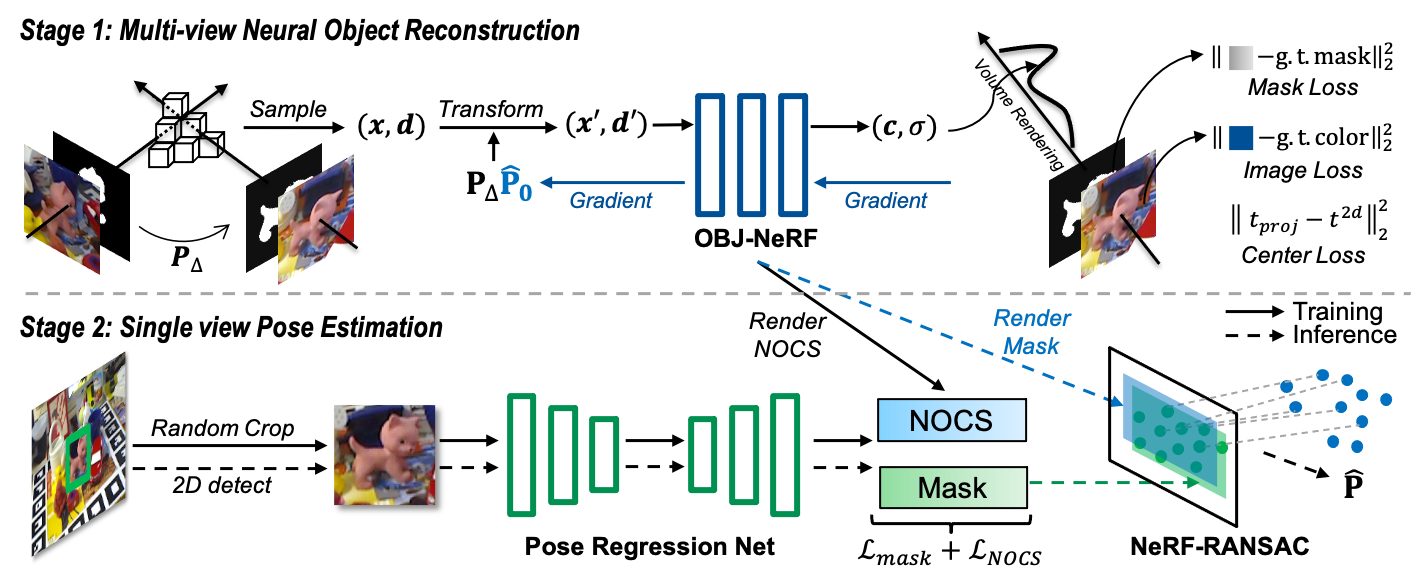}
    \caption{NeRF-Pose Pipeline. {\em Stage 1: Multi-view Neural Object Reconstruction.} During training, we first reconstruct the neural object model from multiple views. The rays generated from different viewpoints are sampled to produce 3D points along the rays (top left).
    The sampled points and directions $(\textbf{x}, \textbf{d})$ are then transformed into an object-centric coordinate system according to the estimated object pose $\hat{\textbf{P}_0}$ and relative camera pose $\mathbf{P}_{\Delta}$.
    The transformed 3D points and directions $(\textbf{x}', \textbf{d}')$ are passed through the learnable NeRF model. We can get the image RGB values and the mask with the NeRF rendering process. The image rendering loss and segmentation mask loss are used to guide the learning of the neural object model.
    {\em Stage 2: Single View Object Pose Estimation}. In stage two, we directly regress the object coordinate map and the segmentation mask supervised by the rendering results from the neural object model.
    At inference, we first detect and crop the objects from 2D images (lower left). The coordinates and segmentation mask are predicted from the pose regression network.
    In the end, the NeRF-enabled PnP+RANSAC is performed to recover the precise object pose in the canonical object space defined at the reconstruction stage.
    }
    \label{fig:pipeline}
\end{figure*}

The first type of pose estimation methods that gained popularity in recent years are dense correspondence-based methods~\cite{zakharov2019dpod,hodan2020epos,wang2019normalized,li2019cdpn,park2019pix2pose,shugurov2021dpodv2,shugurov2021multi}. While being different in implementation, their common denominator is the key idea to train a neural network to predict 2D-3D correspondences between each object pixel in the image and the 3D location of the corresponding point on the object's surface. Those correspondences are consecutively used either with PnP+RANSAC~\cite{lepetit2009epnp,fischler1981random} or the Umeyama algorithm~\cite{umeyama1991least} to compute the 6D object pose.  DPOD~\cite{zakharov2019dpod,shugurov2021dpodv2} is proposed to use discrete UV maps to uniquely parameterize the object surface. With this parameterization, the UNet-like network~\cite{ronneberger2015u} predicts two discrete UV coordinates for each visible pixel occupied by the object. Pix2Pose~\cite{park2019pix2pose} and CDPN~\cite{li2019cdpn} leverage two-stage detectors and use 3D normalized vertex coordinates to parameterize the correspondences. Correspondences in CDPN are only used to estimate rotation, while translation there is predicted directly. NOCS~\cite{wang2019normalized} uses Mask-RCNN~\cite{he2017mask} and a normalized object coordinate space to predict correspondences. However, NOCS mainly focuses on estimating the scale and 6D pose for unseen objects. EPOS~\cite{hodan2020epos} extends the idea of dense correspondences by parameterizing each 2D-3D correspondence with its location within a discrete object fragment. Multi-fragment perspectives and many-to-many 2D-3D correspondences enable it to handle symmetric objects more effectively.

Furthermore, CosyPose~\cite{labbe2020cosypose} and Self6D~\cite{wang2020self6d} use a similar pose parameterization which allows for direct pose prediction. They predict the pose by estimating the 2D center location of the object and its distance to camera center. Combined with the intrinsic parameters of the camera, it gives the estimate of the translation component of the 6D pose. Allocentric rotation parameterization is used for simpler rotation prediction. Self6D is first trained on synthetic data and then fine-tuned on real data without pose annotations in a self-supervised manner. 
AAE~\cite{sundermeyer2018implicit}, leverages manifold learning to retrieve a descriptor of a given object patch from a pre-computed database consisting of descriptors of the same object under various rotation. The translation is estimated based on the object scale in the image. MHP~\cite{manhardt2019explaining} predicts multiple pose hypothesis to estimate the pose of symmetric or occluded objects. 
GDR-Net\cite{wang2021gdr} and SO-Pose\cite{di2021so} use the combination of dense correspondences and direct pose estimation by first predicting the dense correspondences and then performing the learning-based pose prediction, which they name \textit{patch-pnp}. And, SO-Pose introduces self-occlusion information to make the correspondences more stable.

Methods above are all fully supervised with the accurate pose labels and object CAD models available. The following two methods relax the constraints and could be trained without object CAD models. LieNet~\cite{do2019Lienet} directly regresses the pose with the Mask-RCNN~\cite{he2017mask} as backbone network. With known object pose labels and camera intrinsics, Cai et al.~\cite{ianreid2020reconstruct} supervise the coordinate prediction network with the multi-view consistency, minimizing re-projection error across different views. Limited by the accuracy of the coordinate prediction network, the reprojection error is insufficient to guide reliable correspondence learning. LatentFusion\cite{park2020latentfusion} trains the implicit neural object representation, which, at reference, takes the multi-view RGBD images of well-calibrated unseen objects as input and reconstructs the object model. Then, the render-refine network is used to estimate the object pose from RGBD input iteratively. Bundle-SDF~\cite{wen2023bundlesdf} performs 6D tracking and reconstructs an object without assuming a CAD model from a video. WeLSA~\cite{welsa6d} generates pose labels for weakly labeled data by training on very few labeled data using shape alignment and feature alignment. Although WeLSA works with very few labeled data, they employ depth maps for training the pipeline and estimating pose labels for weakly labeled data. 

Concerning weakly supervised methods, there are few works related to ours.
On the basis of NeRF\cite{mildenhall2020nerf}, i-NeRF~\cite{yen2020inerf} presents a differentiable camera pose refinement method, which treats the camera pose as network parameters and iteratively updates the pose by minimizing the discrepancy between the input images and the rendered outputs in particular views. However, because of the high computational burden, it takes about half a minute to process one image and is very sensitive to the pose initialization. BARF~\cite{lin2021barf} and NeRF--\cite{wang2021nerf} introduce the methods to estimate camera poses and train NeRF concurrently. It inspires us to optimize the object pose, while simultaneously reconstructing the object with known relative poses.

Different to the model-free approach~\cite{ianreid2020reconstruct}, which optimizes the network via minimizing the re-projection error, we instead aim at densely generating accurate 2D-3D correspondences between input images and an implicit NeRF object representation obtained from weak labels, yielding better performance.

\section{Methods}\noindent
In this section, we present NeRF-Pose for 6D pose estimation, which only requires weak supervision. We assume that real images with 2D ground truth segmentation masks and relative camera poses are available during training. We first present OBJ-NeRF, an implicit 3D model representation, learned under above-defined constraints. This representation is then used to generate object correspondence maps, which is afterwards used to train our proposed pose regression network. Finally, the regressed correspondences are used in a novel NeRF-enabled PnP+RANSAC algorithm for iterative pose estimation.

\subsection{OBJ-NeRF} 
NeRF~\cite{mildenhall2020nerf} and its followups recover the 3D scene from multiple views with known camera poses. Since we deal with the problem of 6D object pose estimation, our aim is to compute object-centric NeRF and use it as an implicit 3D model representation for object pose estimation. Thus, we propose to modify the original NeRF approach. We take as input the images, segmentation masks and relative camera poses, and output implicit object-specific NeRF representation, named OBJ-NeRF.
As OBJ-NeRF reconstruction relies only on relative camera poses, it will produce a 3D model in some uncertain coordinate systems.
In order to make use of OBJ-NeRF for the purpose of 6D pose estimation based on dense 2D-3D correspondences 
it is necessary to recover it with respect to some reference coordinate systems. 
To achieve this, simultaneously with the NeRF reconstruction, we propose to regress poses of the NeRF reconstructed object with respect to some chosen reference frames. These estimated poses will be used later for the generation of correspondence maps needed for training the correspondence estimation network.

NeRF encodes a 3D scene as a continuous 3D representation using an MLP function $\mathcal{N}_{\psi}: \mathbb{R}^3 \times \mathbb{R}^3 \to \mathbb{R}^3 \times  \mathbb{R}$ with $(\mathbf{x};\mathbf{d}) \mapsto (\mathbf{c}; \sigma)$, which predicts the RGB color $\mathbf{c} \in \mathbb{R}^3$ and volume density $\sigma \in \mathbb{R}$ for each input 3D point $\mathbf{x} \in \mathbb{R}^3$ and its view direction $\mathbf{d} \in \mathbb{R}^3$. It can be summarized as $\mathcal{N}_{\psi}(\mathbf{x},\mathbf{d}) = (\mathbf{c}, \sigma)$, where $\psi$ is the set of network parameters.

Based on the NeRF scene representation, OBJ-NeRF reconstructs the object from multiple images, and optimizes the reference object pose at the same time. It takes images $I$, object masks $\mathbf{M}$, camera intrinsics $\mathbf{K}$, relative poses $\mathbf{P}_{\Delta}$ as training input and outputs the rendered object view $\hat{I}$, its mask $\hat{\mathbf{M}}$, correspondence map $\hat{\mathbf{O}}$ as well as the reference object pose $\mathbf{P}_0$ in respect to the reference image $I_0$. As shown in stage 1 in Fig.~\ref{fig:pipeline}, the steps of the full procedure, from input images to rendering the outputs, can be named as \textit{Sample points}, \textit{Transform points}, \textit{Calculate}, and \textit{Render}. 

\textbf{\textit{Sample points.}} Given 2D pixel coordinates $\mathbf{u} \in \mathbb{R}^2$ in an input image $I$, we can express a 3D point $\mathbf{x}$ along the viewing ray at depth $z$ as $\mathbf{x} = \mathcal{K}^{-1}(\mathbf{u},z,\mathbf{K})$, where $z\in [z_{near}, z_{far}]$ is the depth sampling interval and $\mathcal{K}^{-1}$ is the inverse projection function from image plane to 3D space.  
Different from the scene reconstruction, to reconstruct the object centric model, we limit the ray sampling interval to be close to the estimated object center $\hat{t}^{3d}$, denoted as $(z_{near}, z_{far}) = (|\hat{t^{3d}}|-s, |\hat{t^{3d}}|+s)$, where $|\hat{t^{3d}}|$ is the distance from estimated object center to the camera center, and $s \in \mathbb{R}$ is larger than the object scale. Uniform sampling or weight guided re-sampling can be applied here to generate $N$ points along each ray.

\textbf{\textit{Transform points.}} 
Different from scene-centric BARF~\cite{lin2021barf} and iNeRF~\cite{yen2020inerf}, which optimize absolute camera poses in some arbitrary coordinate systems, we perform object-centric reconstruction, which instead estimates the object poses in the camera coordinate system.  
Therefore, the sampled points $(\mathbf{x},\mathbf{d})$ from camera coordinate system are transformed to the object coordinate system $(\mathbf{x'},\mathbf{d'})$ according to the object pose $\hat{\mathbf{P}_i}$ with respect to the image $I_i$. However, poses of image objects are unavailable and thus need to be estimated.
Notably, owing to our weakly-supervised settings, where relative camera poses from $I_0$ to $I_i$, denoted by $\textbf{P}_{\Delta i0}$, are accessible, only $\mathbf{P}_0$ needs to be optimized. Thus, arbitrary poses can then be computed by $\mathbf{P}_i = \mathbf{P}_{\Delta i0} \mathbf{P}_0$.
We further define the transformation functions: $\mathcal{W}^\mathbf{x}_{\mathbf{P}_0}(\mathbf{x}, \mathbf{P}_{\Delta i0}) \to \mathbf{x'}$ and $\mathcal{W}^\mathbf{d}_{\mathbf{P}_0}(\mathbf{d}, \mathbf{P}_{\Delta i0}) \to \mathbf{d'}$, which transform the coordinate $\mathbf{x}$ and view direction $\mathbf{d}$ into unified object-centric coordinates, with the parameters $\mathbf{P}_0$ to be optimized.

\textbf{\textit{Calculate.}} Until now, we have obtained the transformed ray points and their directions $(\mathbf{x'},\mathbf{d'})$. Then, we feed $(\mathbf{x'},\mathbf{d'})$ into the original NeRF network $\mathcal{N}_{\psi}$ to get color and volume density prediction $(\mathbf{c}, \sigma)$. In summary, our OBJ-NeRF network $\mathcal{N}_{\psi}$ can be formulated as:
\begin{equation}
\small
     \mathcal{N}_{\psi}(\mathbf{x}^{'}, \mathbf{d}^{'}) = \mathcal{N}_{\psi, \mathbf{P}_0}
     (\mathcal{W}^{\mathbf{x}}(\mathbf{x},\mathbf{P}_i), \mathcal{W}^{\mathbf{d}}(\mathbf{x},\mathbf{P}_i)) = (\mathbf{c},\sigma),
\label{eq:nerf}
\end{equation}
where $\psi$ and $\mathbf{P}_0$ are the network parameters. 

\textbf{\textit{Render.}} Akin to NeRF, we use volume rendering approach to map a set of calculated data $(\mathbf{c}, \sigma)$ to the image plane along the rays. The volume rendering function sums up the product of transmittance $T_i$ and alpha value $\alpha_i$ of $N$ sampled points along the ray, which is differentiable. Following \cite{mildenhall2020nerf,yang2021objectnerf}, the rendered color $\mathbf{\hat{C}}$, mask $\mathbf{\hat{M}}$ and coordinates $\mathbf{\hat{O}}$ can be formulated as:
\begin{equation}
\small
    \mathbf{\hat{C}}(\mathbf{u}) = \sum_{i=1}^N T_i \alpha_i\mathbf{c}_i, \mathbf{\hat{M}}(\mathbf{u}) = \sum_{i=1}^N T_i \alpha_i, \mathbf{\hat{O}}(\mathbf{u}) = \sum_{i=1}^N T_i \alpha_i \mathbf{x}_i,
\label{eq:rendering}
\end{equation}
with $T_i=\exp(-\sum_{1\leq j \leq i-1}\sigma_j\delta_j)$, and $\alpha_i=1-\exp(-\sigma_i\delta_i)$, where $\delta_j$ is the sampling distance between sampled adjacent points.

\textbf{\textit{Constrain object pose.}} With only 2D bounding box and segmentation, where object center information is unavailable, the object canonical pose center has no constraint. Accordingly, we constrain the object center by projecting the object center $\hat{t^{3d}}$ to each view.
More specifically, we minimize the reprojection error between the projected object center and 2D bounding box center $t^{2d}$ on the image plane by minimizing $e = ||\mathcal{K}(\mathcal{W}^{x}_{\mathbf{P}_0}( \hat{t^{3d}_0}, \mathbf{P}_{\Delta}), \mathbf{K}) - t^{2d}||^2_2$.

\textbf{\textit{Loss function.}} We leverage the 2D segmentation mask and the segmented image as the supervision signals for the rendered pixel mask and color value. 
Denoting the segmentation mask at pixel $\mathbf{u}$  as $\mathbf{M}(\mathbf{u}) \in \{0, 1\}$, and the color value as $\mathbf{C}(\mathbf{u}) \in \mathbb{R}^3$, our loss function for the reconstruction stage can be defined as:
\begin{equation}
\small
    \begin{split}
        \mathcal{L}_{rec} = & \sum_{i}( \sum_{\mathbf{u}} \mathbf{M}(\mathbf{u}) || \mathbf{C}(\mathbf{u}) - \mathbf{\hat{C}}(\mathbf{u}) ||^2_2\\
                          &+ \sum_{\mathbf{u}} ||\mathbf{M}(\mathbf{u}) - \mathbf{\hat{M}}(\mathbf{u})||^2_2\\
                          &+ ||\mathcal{K}(\mathcal{W}^{\mathbf{x}}_{\mathbf{P}_0}( \hat{t^{3d}_0}, \mathbf{P}_{\Delta i0}), \mathbf{K}) - t^{2d}_i||^2_2 ),
    \end{split}
\end{equation}
where $i$ iterates the training image views, and $\textbf{u}$ iterates the image pixels.

In the end, the newly defined object canonical pose is determined by optimizing the object pose in multi-view settings. With the learned canonical pose and neural model, the \textit{NOCS} map ground truth can be rendered for the next pose estimation stage.

\begin{figure}
    \centering
    \includegraphics[width=\linewidth]{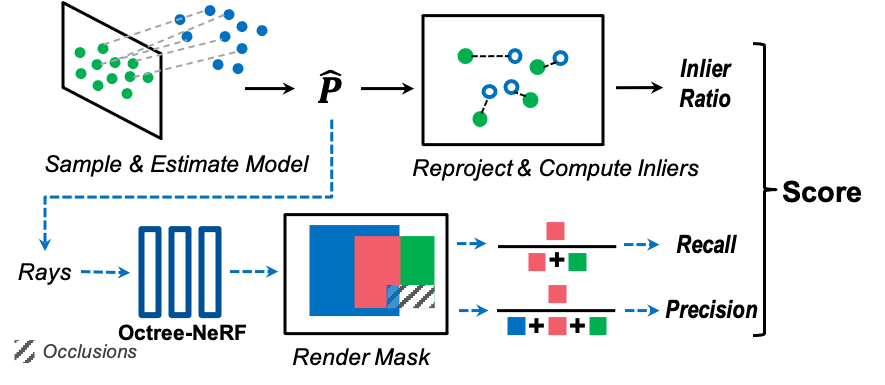}
    \caption{Illustration of a single iteration of the NeRF-enabled RANSAC. Similar to a standard PnP+RANSAC procedure~(upper top), the pose hypothesis is first calculated by the PnP algorithm from sampled 2D-3D correspondences. Scored by Inlier Ratio, the pose hypothesis with most inliers is selected as the estimated results.
    Different to standard PnP+RANSAC, we further assemble Recall and Precision into our criteria of pose selection~(bottom). Detailly, we first render the object mask in this pose hypothesis using our well-trained octree-NeRF. Then, we calculate Precision by the Intersection (red)-over-Union (blue+red+green) between the rendered mask and the mask from our pose regression network. Recall is defined as the ratio of the intersection (red) over the area of regressed mask (red+green). In the end, the selection score are calculated as the weighted sum of Inlier Ratio, Precision, and Recall.
    }
    \label{fig:ransac}
\end{figure}

\begin{figure*}
    \centering
    \includegraphics[width=\linewidth]{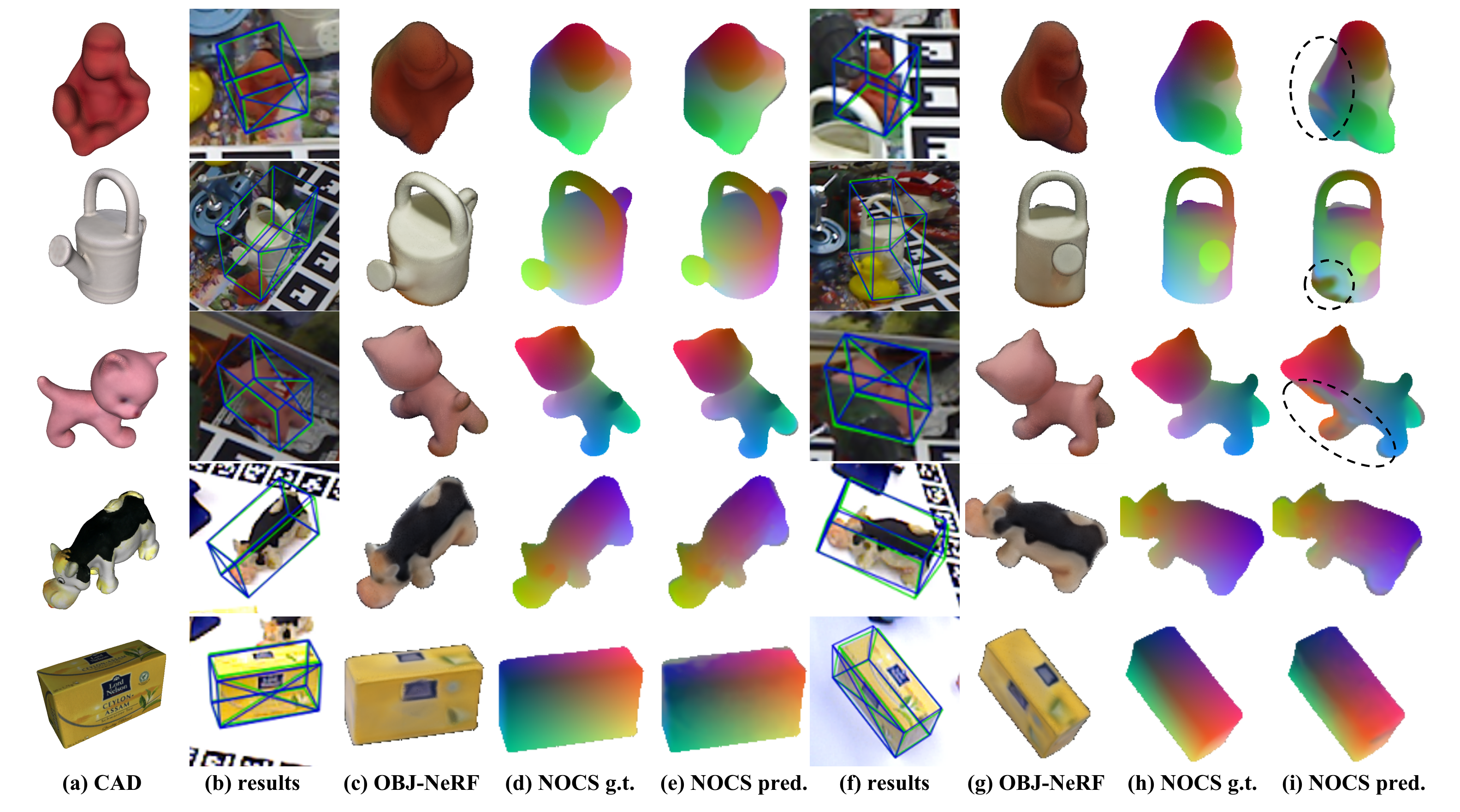}
    \caption{Qualitative results on \textbf{LM} and \textbf{HBD}. 
    Column (a) shows the CAD models used as reference ground truth shapes.  
    Column (b) and (f) illustrate the predicted object poses (blue) and ground truth poses (green).
    Column (c) and (g) are the rendered images from our well-trained OBJ-NeRF model,
    Column (d) and (h) are the produced \textit{NOCS} map from our OBJ-NeRF, which are also \textit{g.t.} of regression network.
    Column (e) and (i) are the predicted \textit{NOCS} from our regression network. 
    Black dash circles show the \textit{NOCS} prediction of occlusion regions.
    }
    \label{fig:resvis}
\end{figure*}

\subsection{Pose Estimation}\label{sec:pose}

The overall three-step pipeline is shown in Stage 2 in Fig.~\ref{fig:pipeline}. The first two steps depend on separately trained convolutional neural networks. The third step is purely optimization-based and does not require training. The first step represents the off-the-shelf 2D detector trained on ground truth crops of the objects of interest. In practice, we use YOLOv3~\cite{farhadi2018yolov3}. 
In the second step, a pose regression network is trained on images $I$ to predict object coordinates $\mathbf{O}$ and segmentation mask $\mathbf{M}$. The third step is our NeRF-enabled PnP+RANSAC algorithm, to improve the performance of pose calculation by introducing the NeRF-Mask renderer.

\textbf{\textit{Coordinates regression.}} Our pose regression network is inspired by DPoD~\cite{zakharov2019dpod} and CDPN~\cite{li2019cdpn}, the state-of-the-art dense correspondence-based methods for indirect pose estimation. We use ResNet\cite{he2016resnet} as encoder backbone, and the decoder contains four upsampling layers. Our pose regression network outputs the predicted segmentation $\hat{\mathbf{M}}$ and the object coordinates $\hat{\mathbf{O}}$, which encodes correspondences between input image pixels and the OBJ-NeRF 3D model representation.   

The loss function is defined similarly to other depth regression works~\cite{barron2019general, hu2019revisiting}. Our loss function is composed of the segmentation loss and the normalized coordinate regression loss:
\begin{equation}
\small
    \mathcal{L} = w_1 \mathcal{L}_{M} + w_2 \mathcal{L}_{NOCS}.
\end{equation}
The $\mathcal{L}_M$ loss is defined by the mean squared error (MSE) between predicted mask $\mathbf{\hat{M}}$ and ground truth mask $\mathbf{M}$. $\mathcal{L}_{NOCS}$ is defined between the predicted object \textit{NOCS} map $\hat{\mathbf{O}}$ and their ground truth $\mathbf{O}$ rendered from the implicit neural network OBJ-NeRF as given below:
\begin{equation}
    \small
    \mathcal{L}_{NOCS} = \lambda_{1} \mathcal{L}_{mse} + \lambda_{2} \mathcal{L}_{grad} + \lambda_{3} \mathcal{L}_{normal},
\end{equation}

The first term $\mathcal{L}_{mse}$ is the element-wise loss that measures the average distance between these two coordinates written as:
\begin{equation}
    \small 
    \mathcal{L}_d = \| \mathbf{M} \odot (\mathbf{O} - \mathbf{\hat{O}}) \|^2_2.
\end{equation}

$\mathcal{L}_{grad}$ and $\mathcal{L}_{normal}$ are used to penalize the coordinate errors in first and second order. Similar to the loss of~\cite{hu2019revisiting}, they are defined as: 
\begin{equation}
    \scalemath{0.8}{
    \mathcal{L}_{grad} =  \| \mathbf{M} \odot (\nabla_{x} \mathbf{\hat{O}} - \nabla_{x}\mathbf{O})\|^2_2 + \| \mathbf{M} \odot (\nabla_{y} \mathbf{\hat{O}} - \nabla_{y} \mathbf{O}) \|^2_2},
\end{equation}
\begin{equation}
    \small
    \mathcal{L}_{normal} = \| \mathbf{M} \odot (1 - \cos(n(\mathbf{\hat{O}}), n(\mathbf{O})) \|^2_2,
\end{equation}
where $\nabla_{x}$ and $\nabla_{y}$ are the gradients for the coordinates along $x$ and $y$ axis, $n(o) = [\nabla_{x}(o), \nabla_{y}(o), -1]$ 
defines the surface normal vector and $\cos(n_1, n_2)$ measures the cosine similarity of the two vectors.  

\textbf{NeRF-enabled PnP+RANSAC.} With the predicted dense 2D-3D correspondences, PnP+RANSAC-based method~\cite{zakharov2019dpod, li2019cdpn,hodan2020epos} are typically used for estimating object pose $\hat{\mathbf{P}}$. 
As shown in Fig.~\ref{fig:ransac}, the PnP+RANSAC~\cite{fischler1981random} algorithm iteratively selects a minimum number of correspondences for pose estimation and calculates object pose using PnP method. The pose hypothesis supporting the most inliers is selected as the calculated pose results. 

The criteria of pose selection in PnP+RANSAC is inlier ratio $S_{IR}$, which highly depends on the correspondence quality. To make PnP+RANSAC more stable, we add two extra pose selection criteria which are independent to the correspondences.
As shown in Fig.~\ref{fig:ransac}, with the calculated pose hypothesis and our implicit object representation, we can render the object mask $\hat{\textbf{M}}_{nerf}$ on the image plane without occlusion. 
With the mask $\hat{\textbf{M}}_{pose}$ predicted from our pose regression network, we add the \textit{Overlap Recall} $S_{Recall}$ and \textit{Overlap Precision} $S_{Prec}$ to the score which is defined as: $S_{Recall} = |\hat{\textbf{M}}_{nerf} \bigcap \hat{\textbf{M}}_{pose}| / | \hat{\textbf{M}}_{pose} | $ and $S_{Prec} = |\hat{\textbf{M}}_{nerf} \bigcap \hat{\textbf{M}}_{pose}| / |\hat{\textbf{M}}_{pose} \bigcup  \hat{\textbf{M}}_{nerf}|$.
The final score is defined as $S = v_1 S_{IR} + v_2S_{Prec} + v_3 S_{Recall}$.

Following Plenoctree~\cite{yu2021plenoctrees}, we convert the well-trained OBJ-NeRF model to an octree-based data structure, namely \textit{octree-NeRF}, which stores the average value of $\mathbf{c}, \sigma$ sampled in the leaf voxel of octree. The NeRF calculation is simplified by the octree indexing, which significantly accelerates the mask rendering.
\section{Experiments}

\begin{table*}[h]
    \begin{center}
    \small
    \tabcolsep0.06in 
    \caption{\textbf{LM} results in on ADD-10 metric. *denotes that the objects is symmetric and is evaluated in ADD-S. \textbf{Our-pose} denotes our results trained on 6D pose labels, and \textbf{Our-weak} denotes training on camera relative pose labels. }
    \begin{tabular}{c | c c c c c | c c c c c c c c}
    \bottomrule
        Object  & DPOD & PVNet & CDPN & GDR & SO-Pose & LieNet & Cai. & \textbf{Ours-sam} & \textbf{Ours-pose}&  \textbf{Our-pose} & \textbf{Our-weak} \\ \hline
        &  \cite{zakharov2019dpod}& \cite{peng2019pvnet} & \cite{li2019cdpn} & \cite{wang2021gdr} & \cite{di2021so} & \cite{do2019Lienet} & \cite{ianreid2020reconstruct}& \textit{w/o NeRF} & \textit{w/o NeRF}& \textit{w/ NeRF} & \textit{w/ NeRF}  \\
        \hline
\textit{CAD} &  \multicolumn{5}{c|}{\textit{w/ CAD}} & \multicolumn{6}{c}{\textit{w/o CAD}} \\
\hline
Ape    & 53.3 & 43.6 & 64.4 & - & - & 38.8 & 52.9 &  50.1 & 69.4 & 89.1 & \textbf{93.1} \\
Bvise   & 95.2 & 99.9 & 97.8 & - & - & 71.2 & 96.5 &  99.4 & 99.4 & 99.3 & \textbf{99.6} \\
Cam     & 90.0 & 86.9 & 91.7 & - & - & 52.5 & 87.8 & 97.7 & 98.3  & 98.7 & \textbf{98.9} \\
Can     & 94.1 & 95.5 & 95.9 & - & - & 86.1 & 86.8 & 98.7 & 97.8 & 99.1 & \textbf{99.7} \\
Cat      & 60.4 & 79.3 & 83.8 & - & - & 66.2 & 67.3 &77.2 & 77.8 & 97.1 & \textbf{98.1} \\  
Drill   & 97.4 & 96.4 & 96.2 & - & - & 82.3 & 88.7 & 99.1 & 99.6 & 97.4 & \textbf{98.7} \\
Duck    & 66.0 & 52.6 & 66.8 & - & - & 32.5 & 54.7 & 57.4 & 69.7 & 90.3 & \textbf{94.2} \\
Eggbox* & 99.6 & 99.2 & 99.7 & - & - & 79.4 & 94.7 & 89.1 & 99.9& 99.6 & \textbf{99.9} \\
Glue*    & 93.8 & 95.7 & \textbf{99.6} & - & - & 63.7 & 91.9 & 100 & 98.9 & 98.1 & 99.3 \\
Holep   & 64.9 & 81.9 & 85.8 & - & - & 56.4 & 75.4 & 90.3 & 89.4 &94.3 & \textbf{96.5} \\
Iron.   & 99.8 & 98.9 & 97.9 & - & - & 65.1 & 94.5 & 100 & 99.8 & \textbf{98.1} & 97.8 \\
Lamp    1 & 88.1 & 99.3 & 97.9 & - & - & 89.4 & 96.6 & 98.7 & 99.8 & 97.9 & \textbf{98.7} \\
Phone    & 71.4 & 92.4 & 90.8 & - & - & 65.0 & 89.2 & 90.2 & 94.8 & 96.4 & \textbf{97.3} \\
\hline
Mean   & 82.6 & 86.3 & 89.9 & 93.7 & 96.0 & 65.2 & 82.9 & 88.3($\uparrow5.4$)&91.8($\uparrow8.9$) &  96.6($\uparrow13.7$) & \textbf{97.8}($\uparrow14.9$)\\\bottomrule
\end{tabular}
\label{tab:linemod}
\end{center}
\end{table*}

\subsection{Datasets}\noindent
In this section, we conduct extensive experiments to demonstrate that our proposed weakly-supervised pose estimation method, \textbf{NeRF-Pose}, produces highly-accurate 6D object pose estimation.
 
\textbf{Datasets.} We conduct our experiments on three publicly available datasets: Linemod~(LM)\cite{hinterstoisser2012model}, Linemod-Occlusion~(LMO)\cite{brachmann2014learning}, T-Less\cite{hodan2017tless} and HomebrewedDB~(HBD)\cite{kaskman2019homebreweddb}.
The LM dataset is a standard benchmark for 6D object pose estimation of textureless objects. It offers 13 objects of various sizes in the scenes with large background clutter but with almost no occlusion. 
The LMO dataset consists of 8 objects from LM dataset but provides more challenging test data with more occlusion. 
According to our problem settings, we train our model on the real data in LM and strictly follow the training/testing split proposed in \cite{brachmann2014learning}.
On the LMO dataset, to make a fair comparison to Cai et al.~\cite{ianreid2020reconstruct}, we train our method with the real training data of LM.
Since the HBD dataset has no real training data, we provide an \textbf{R}eal data \textbf{EXT}tension of the HBD dataset, named HBD-REXT. It consists of real images of each HBD object presented in BOP~\cite{hodan2018bop} challenge. These images are captured by the Azure Kinect camera with only relative poses and segmentation mask provided. We will make the HBD-REXT dataset publicly available soon.

\textbf{Evaluation Metrics.}
On the LM and LMO datasets, we report the standard ADD(-S) metrics~\cite{hinterstoisser2012model} with the 10\% diameter threshold, as it is the most prevalent pose quality metric for these two datasets. The $n^{\circ}, n~cm$ metric~\cite{shotton2013scene} measures whether the rotation error is less than $n^\circ$ and the translation error is below $n$ cm. Besides, we also use $n^{\circ}$ and $n~cm$ ~\cite{shotton2013scene}, which measure whether the rotation error is less than $n^\circ$ and the translation error is below $n$ cm, respectively. 
Moreover, following the setup of the BOP challenge, which aims at unifying the evaluation of 6D pose estimation methods, we report the following metrics for the HBD dataset: Visible Surface Discrepancy (VSD)~\cite{hodan2018bop,hodavn2016evaluation}, Maximum Symmetry-Aware Surface Distance (MSSD)~\cite{drost2017introducing}, Maximum Symmetry-Aware Projection Distance (MSPD) and Average Recall, which is computed as: $AR = (AR_{VSD} + AR_{MSSD} + AR_{MSPD} )/3$. 

Since we predict the poses in a canonical orientation, it is necessary to transform them to absolute poses. This is done by transforming them with the offset poses~(obtained from the \textit{g.t.} poses), which represent the transformation from absolute poses to the canonical poses. Note that this transformation is only leveraged for evaluation.

\begin{table*}[h]
    \begin{center}
    \small
    \tabcolsep0.03in 
    \caption{Comparisons with state-of-the-art methods on LMO. We report the Average Recall(\%) of ADD(-S) without refinement. \textit{real} denotes the same real data as LM. \textit{syn} denotes self-generated synthetic data, and \textit{pbr} denotes blender rendered synthetic data from BOP\cite{bopchallenge}. * denotes the symmetric objects. \textbf{Our-pose} denotes our results with accurate pose labels and \textbf{Our-weak} is with relative pose labels. \textit{w/o NeRF} denotes our results using original PnP+RANSAC and \textit{w/ NeRF} is our method with our NeRF-enabled PnP+RANSAC. 
    }  
    \begin{tabular}{c | c c c c c c | c  c | c c c | c c}
\bottomrule
& PoseCNN  & PVNet & Single-Stage & HybridPose & GDR &  SO-Net & GDR & SO-Net  & Cai. & \textbf{Our-pose} & \textbf{Our-pose} & \textbf{Our-weak} & \textbf{Our-weak} \\
& \cite{xiang2017posecnn} & \cite{peng2019pvnet} & \cite{hu2020single} & \cite{song2020hybridpose} & \cite{wang2021gdr} & \cite{di2021so} & \cite{wang2021gdr} & \cite{di2021so} & \cite{ianreid2020reconstruct} & \textit{w/o NeRF} & \textit{w/ NeRF} & \textit{w/o NeRF} & \textit{w/ NeRF}\\ 
\hline
\textit{CAD} & \multicolumn{6}{c|}{\textit{w/ CAD}} & \multicolumn{2}{c|}{\textit{w/ CAD}} & \multicolumn{3}{c|}{\textit{w/o CAD}} & \multicolumn{2}{c}{\textit{w/o CAD}} \\
\hline
training & \multicolumn{6}{c|}{\textit{real+syn}} &  \multicolumn{2}{c|}{\textit{real+pbr}} &  \multicolumn{3}{c|}{\textit{real}}  &  \multicolumn{2}{c}{\textit{real}} \\
\hline
Ape     &  9.6     & 15.8  & 19.2         & 20.9        & 39.3    &  46.3   & 46.8  & 48.4    & 7.10  & 46.8  & 46.9  & 48.3 &\textbf{49.7}   \\
Can     &  45.2    & 63.3  & 65.1         & 75.3        & 79.2    &  81.1   & \textbf{90.8}  & 85.8    & 40.6  & 79.1  & 86.2  & 81.4 & 86.4   \\
Cat     &  0.9     & 16.7  & 18.9         & 24.9        & 23.5    &  18.7   & \textbf{40.5}  & 32.7    & 15.6  & 20.7  & 27.1  & 28.8 & 26.9   \\
Driller &  41.4    & 65.7  & 69.0         & 70.2        & 71.3    &  71.3   & \textbf{82.6}  & 77.4    & 43.9  & 58.9  & 65.8  & 60.4 & 66.2   \\
Duck    &  19.6    & 25.2  & 25.3         & 27.9        & 44.4    &  43.9   & 46.9  & \textbf{48.9}    & 12.9  & 25.3  & 29.9  & 32.8 & 36.9   \\
Eggbox  &  22.0    & 50.2  & 52.0         & 52.4        & \textbf{58.2}    &  46.6   & 54.2  & 52.4    & 46.4  & 19.6  & 24.9  & 22.8 & 24.4   \\
Glue    &  38.5    & 49.6  & 51.4         & 53.8        & 49.3    &  63.3   & 75.8  & \textbf{78.3}    & 51.7  & 61.0  & 66.3  & 69.8 & 70.9   \\
Holep.  &  22.1    & 36.1  & 45.6         & 54.2        & 58.7    &  62.9   & 60.1  & \textbf{75.3}    & 24.5  & 41.0  & 46.4  & 41.8 & 49.8   \\
\hline
Mean    &  24.9    & 40.8  & 43.3         & 47.5        & 53.0    &  54.3   & 62.2  & \textbf{62.3}    & 30.3  & 44.1  & 49.2  & 48.2 & 51.4($\uparrow21.1$)   \\
    \bottomrule
    \end{tabular}
    \label{tab:lmo}
    \end{center}
\end{table*}

\subsection{Comparisons to the State-of-the-art}\noindent

\textbf{LM.}
We train our method on real images, strictly following the training/testing split from \cite{brachmann2014learning}. 
In order to make a fair comparison to Cai at al.~\cite{ianreid2020reconstruct}, apart from training using relative poses(\textbf{Our-weak}), we also train our method using pose labels(\textbf{Our-pose}). 
In Tab.~\ref{tab:linemod}, our method outperforms the method of Cai at al.~\cite{ianreid2020reconstruct} by a large margin ($\approx14pp$) on a ADD(-S) metric. 
Moreover, our method with weakly-supervised settings performs on-par with the SoTA fully-supervised methods. We also perform an ablation, (\textbf{Ours-sam})  by using SegmentAnything\cite{kirillov2023segment} to generate segmentation masks using ground truth bounding boxes as input instead of using ground truth segmentation masks. The mild accuracy drop(3.5\%) indicates that our approach can be applied in the real world much more easily compared to other approaches using relative poses from the sensor or SFM and segmentation masks obtained using SAM.   

\textbf{LMO.}
We compare our method with SoTA methods in terms of ADD(-S) on LMO. 
Compared to Cai et al.~\cite{ianreid2020reconstruct} trained with pose labels but without CAD models on real images, our method on the same settings achieves 49.2\% on mean ADD score, surpassing it with a large margin of $20pp$. 
In our setting, the CAD models are not accessible, so we do not train our method in self-generated synthetic images or \textit{pbr} images pulished in \cite{hodan2018bop,bopchallenge}.
It is proved that training with more synthetic images can improve the model performance and \textit{pbr} images can improve more. 
So, it is reasonable to claim that we stays comparable with the SoTA fully-supervised methods (GDR-Net~\cite{wang2021gdr}: 53.0\% and SO-Pose~\cite{di2021so}: 54.3\%)  trained on real and author-generated synthetic images.

In Tab.\ref{tab:linemod} and Tab.\ref{tab:lmo}, \textbf{Our-weak}(uses OBJ-NeRF with relative poses) performs better than \textbf{Our-pose} (uses OBJ-NeRF with absolute poses). This is caused by noisy labels in LM, which harm more \textbf{Our-pose} than \textbf{Our-weak}. This is because in \textbf{Our-weak} we optimizes all pose labels by minimizing the reprojection and rendering errors constrained with relative poses. In \textbf{Our-pose} with absolute poses we cannot refine them, since we cannot guarantee to obtain the correct object's scale.


\textbf{HBD-REXT.} To support weakly-supervised training on real images, we extend the HomebrewedDB training set by capturing more real sequences for all objects used in BOP~\cite{hodan2018bop} challenge. We provide about 300 real images for each object with segmentation masks and relative camera poses generated from the markerboard. We train our model on this newly captured data. Since no other method was trained only with relative poses and 2D segmentation, we cannot run other methods using this new weakly-supervised data. In Tab.~\ref{tab:hbd} we report the AR of VSD, MSSD, MSPD metrics on the BOP challenge test set. It illustrates that we stay on-par with the methods trained on synthetic images ~\cite{zakharov2019dpod,hu2020single,li2019cdpn} and fall behind the methods trained on \textit{pbr} images~\cite{hodan2020epos,labbe2020cosypose,li2019cdpn}. Notably, neither CAD models nor ground truth poses are used in the \textbf{our-weak} case, whose results are shown in the last column of Tab.~\ref{tab:hbd}.

\begin{table}[h]
    \begin{center}
    \small
    \tabcolsep0.03in 
    \caption{Comparisons with state-of-the-art methods on \textbf{HBD}.}
    \resizebox{0.48\textwidth}{!}{
    \begin{tabular}{c | c c c | c c c c | c}
\bottomrule
      & DPoD  & Single & CDPN & Pixel2Pose & EPOS  & CosyPose & CDPN  & \textbf{Our-weak}  \\
      & \cite{zakharov2019dpod} & \cite{hu2020single} & \cite{li2019cdpn}
& \cite{park2019pix2pose} & \cite{hodan2020epos} & \cite{labbe2020cosypose} & \cite{li2019cdpn} & \textit{w/ NeRF} \\
\hline
training & \textit{syn} & \textit{syn} & \textit{syn} & \textit{pbr} & \textit{pbr} & \textit{pbr} & \textit{pbr} & \textit{real-ext} \\
\hline
VSD   & 21.8 & 24.1  & 39.1 & 35.2       & 48.4 & 61.3    & 61.4 & 44.0 \\
MSSD  & 26.2 & 25.0  & 45.1 & 39.4       & 52.7 & 63.4    & 70.8 & 48.6 \\
MSPD  & 37.9 & 38.8  & 56.9 & 59.4       & 72.9 & 72.1    & 84.5 & 62.8 \\
\hline
AR    & 28.6 & 29.3  & 47.0 & 44.6       & 58.0 & 65.6    & 72.2 & 51.8 \\
\hline
Time(s)  & 0.18 & 0.19  & 0.31 & 0.98       & 0.66 & 0.42    & 0.27 & 0.25 \\
    \bottomrule
    \end{tabular}
    }
    \label{tab:hbd}
    \end{center}
\end{table}

 \textbf{T-Less.} We evaluate our pipeline on T-Less dataset. The T-Less dataset comprises 30 objects with real training images. We train our model using relative camera poses and real training images. In Tab.~\ref{tab:tless} we report the AR of VSD, MSSD, MSPD metrics on the BOP challenge test set. We achieve closer to benchmark accuracy despite not using a CAD model. It shows that Nerf can learn accurate geometry and render correspondences which are usually extracted from the CAD model. SurfEmb performs better than our approach as their approach is tailored for symmetric objects and also employs an inference pipeline with 2.2s. However, the results compared to other regression-based, Dpod and DpodV2, show that our approach can perform equally better employing NeRF.  

\begin{table}[h]\centering
\scriptsize
\caption{Comparisons with state-of-the-art methods on T-Less. We report the VSD, MSPD, MSSD, AR metrics as described in the BOP challenge without refinement. CAD refers to  the approaches assuming that the CAD model is available for training}  
\begin{tabular}{l|rrrr|rrrr}\toprule
Approach &Dv2 &SurfEmb &EP &CP &Dv2 &CDPN &Ours \\\midrule
&\cite{shugurov2021dpodv2} & \cite{haugaard2022surfemb} & \cite{hodan2020epos} & \cite{labbe2020cosypose} & \cite{shugurov2021dpodv2} & \cite{li2019cdpn}\\
\midrule
CAD &Y &Y &Y &Y &N &N &N \\ \midrule
VSD &0.57 &0.5 & &0.57 &0.46 &0.49 &0.45 \\
MSSD &0.62 &0.53 & &0.59 &0.49 &0.67 &0.49 \\
MSPD &0.76 &0.83 &0.63 &0.76 &0.59 &0.41 &0.66 \\\midrule
AR &0.65 &0.62 &0.47 &0.64 &0.51 &0.37 &0.54 \\
\bottomrule
\end{tabular}
\label{tab:tless}
\end{table}

\subsection{Ablations}
\textbf{Shape Analysis.} The visual results of the objects represented in OBJ-NeRF can be found in Fig.~\ref{fig:resvis}(c, g).
Compared to their reference CAD models (Fig~\ref{fig:resvis}(a)), our OBJ-NeRF keeps both the detailed shapes and the texture information. Accurate shapes obtained in our model promises the accuracy of our consecutive pose estimation. We also observe some minor defects on our predicted shapes, e.g. details of cat eyes and cow legs, which is mainly caused by the imprecise relative pose annotations and imperfect segmentation masks. 

\textbf{View Numbers.} To figure out how the number of views used in NeRF reconstruction affects the pose estimation, we train OBJ-NeRF with 4 different numbers of views---32(B1), 64(B2), 128(B3) and 156(Baseline). 
From the results in Tab.~\ref{tab:ablation}, training using 156 views is slightly better than the others. It actually shows small difference (around $0.5\%$) on ADD metric and almost no influence of the number of views to the results. Thus, we use 156 views for training OBJ-NeRF. Moreover, the results also illustrate the number of views does not bring significant influence on the performance, i.e., the OBJ-NeRF can recover the highly precise object model, though trained with fewer views. 
It is an interesting conclusion that can inspire us to use few-shot labels for reconstruction and pose estimation. 

\textbf{Regression Loss.} 
We study the contribution of our pose regression loss. As shown in Tab.~\ref{tab:ablation} C1-C3, the loss with gradient and normal loss has the best performance, which indicates the efficiency of our utilized loss functions.

\begin{figure}
   \centering
    \includegraphics[width=\linewidth]{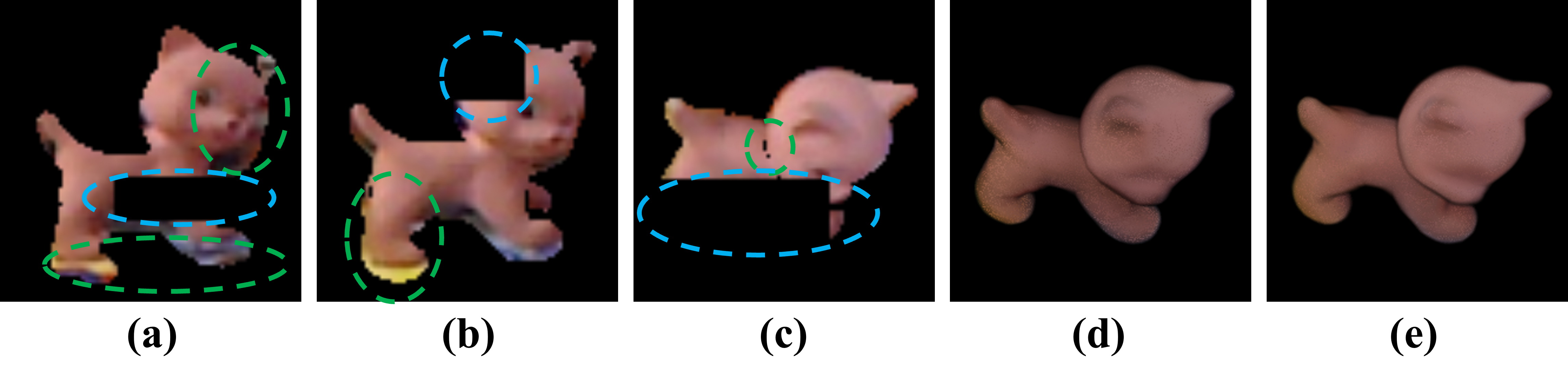}
        \caption{(a),(b) and (c) are masked RGB images (green ellipses are noises and blue ellipses are our generated occlusions), and (d) is rendered with occluded images training, (e) without.  
        }
    \label{fig:noiseocc}
\end{figure}

\textbf{Training with occlusions.} 
To further evaluate the influence of occlusions on the performance of OBJ-NeRF, we add random rectangles onto RGB images to mask out certain areas and simulate occlusions and errors in segmentation masks.
We test the performance on the same objects as in other ablation studies. 
The reconstruction computed from the distorted images (Fig.\ref{fig:noiseocc}(d)) has no clear difference to the non-distorted one (Fig.\ref{fig:noiseocc}(e)).
However, in Tab~\ref{tab:ablation}, training with occlusion(D1) has a slight drop (about $5~pp$ on ADD score) on pose estimation performance. We consider it acceptable, as the occlusion increases the difficulty of the reconstruction. 
Moreover, as indicated by the green ellipses in Fig.\ref{fig:noiseocc}, the \textit{g.t.} masks and pose labels from the LM and HBD are actually noisy. Due to these imprecision, our reconstructions are missing some details. However, all our experiments demonstrate that such reconstructed implicit object models and small pose errors can be tolerated.

\begin{table}[h]
    \begin{center}
    \tabcolsep0.06in 
    \caption{Ablation study on LM subset. We train the whole pipeline for the Ape, the Cat, and the Duck objects. A0 is the baseline setting in our paper, and B1-B3 show the results on different numbers of training views in OBJ-NeRF. C1-C3 are the results using different loss combination. D1 is the result training with generated occlusions. E1-E3 consists of the results from different the PnP+RANSAC strategies.
    }
    \small
    \resizebox{0.48\textwidth}{!}{
    \begin{tabular}{c | l | c | c | c | c }
    \bottomrule
        Case & Method & ADD(0.01d)& 2cm & $2^o$,2cm & $2^o$ \\
        \hline
        A0 & NeRF-Pose & 95.1	& 97.5 & 82.3 & 83.2 \\
        \hline
        B1 & rec. 32 views & 94.6 & 97.0 & 79.6 & 79.9\\
        B2 & rec. 64 views & 94.0 & 96.6 & 78.4 & 78.9\\
        B3 & rec. 128 views & 93.6 & 96.2 & 78.5 & 79\\
        \hline
        C1 & w/o $\mathcal{L}_{grad}$ & 91.5 & 95.2 & 77.2 & 78.4\\
        C2 & w/o $\mathcal{L}_{norm}$ & 91.0 & 94.4 & 76.9 & 78.0\\
        C3 & w/o $\mathcal{L}_{grad}, \mathcal{L}_{grad}$ & 92.9 & 96.1 & 77.7 & 78.3\\
        \hline
        D1 & rec. \textit{w/} occlusion & 89.8($\downarrow 5.3~pp$)	& 93.8 & 74.4 & 76.0 \\
        \hline
        E1 & PnP+RANSAC & 93.6($\downarrow 1.5~pp$) & 96.8 & 83.0 & 83.8 \\
        E2(LMO) & NeRF-PnP+RANSAC  & 51.4 & 48.4 & 9.6 & 12.1 \\
        E3(LMO) & PnP+RANSAC & 48.2 ($\downarrow 3.2~pp$) & 45.7 & 8.7 & 11.6  \\
    \bottomrule
    \end{tabular}
    \label{tab:ablation}
    }
    \end{center}
\end{table}

\textbf{NeRF-enabled PnP+RANSAC.} 
Furthermore, we evaluate the effectiveness of our proposed correspondence solver. We present the results in Tab.~\ref{tab:ablation}, which shows about 2\% improvement on LM dataset on ADD score (E1). Though on $2^\circ$ metric, original PnP+RANSAC is slightly better than ours, our method outperforms it on $2$ cm metric. On ADD score, our method shows superiority over the original PnP+RANSAC. The main reason is that our NeRF-enabled RANSAC incorporates mask-based Recall and Precision in scoring pose hypothesis, as illustrated in Fig.~\ref{fig:ransac}. Since Recall and Precision are more sensitive to translation, our NeRF enabled RANSAC prefers pose hypothesis with less translation error, leading a better performance on both ADD and 2 cm metrics that reflect translation performance.
In Tab.~\ref{tab:ablation} E2-E3, we also report the ADD(-S) score on LMO dataset. We observe an average 3\% improvement using our proposed NeRF-enabled method. The results show the superiority of our proposed NeRF-enabled PnP-RANSAC algorithm over the original one. LMO dataset contains more occlusion data than LM dataset. The occlusions can make mask scores used in NeRF-PnP-Ransac imprecise. In that case overlap recall and precision can deteriorate. 
However, since we weight the inlier score the most with $0.6$ vs. $0.2$ for precision and recall scores, and since our inliers are quite accurate we still get better performance than standard PnP-RANSAC as shown in the Tab.\ref{tab:ablation} for the LM-O benchmark. 

\subsection{Implementation and Runtime Analysis}
We implement our object-centric NeRF network based on the original version of NeRF in\cite{mildenhall2020nerf}, and train the network from scratch. 
For the 2D detector, we use the standard YOLOv3~\cite{farhadi2018yolov3} detector in stage two. An ImageNet~\cite{krizhevsky2012imagenet} pre-trained ResNet34~\cite{he2016resnet} network is leveraged as the backbone of our pose regression network. All networks are trained until convergence.

The training is done on a machine with a Titan RTX GPU with 24GB Memory, an Intel(R) i7-8700K CPU and 24GB RAM. During inference, for a single image with $640\times480$ resolution,  our approach takes about 0.25s for one object, including about 0.03s for YOLOV3 2D detector, 0.01s for pose regression and 0.21s for our pose solver.  

\section{Limitations}
Though we present \textbf{NeRF-Pose} in a weakly-supervised way, considering the training difference in OBJ-NeRF network and our pose regression net, failing to enable end-to-end optimization sometimes leads to local minima. 

As indicated in Tab.~\ref{tab:lmo}, when trained on \textit{pbr} images rendered by \textit{Blender} with high quality from BOP~\cite{hodan2018bop, bopchallenge},
GDR~\cite{wang2021gdr} and SO-Pose\cite{di2021so} gain about 10\% improvement on ADD(-S) metric. Those fully-supervised methods benefit from $pbr$ images that cover more poses and have more realistic occlusion under various light conditions. It inspires us to generate more synthetic training data using our well-trained OBJ-NeRF for better performance.

\section{Conclusion}\noindent
In this paper, we propose \textbf{NeRF-Pose}, a first-reconstruct-then-regress approach for weakly-supervised object pose estimation. 
\textbf{NeRF-Pose} first implicitly reconstructs the object as the proposed neural network, namely OBJ-NeRF, from the weak labels and generates the signals to supervise the correspondences predicted from our pose regression network. At inference, a NeRF-enabled PnP+RANSAC algorithm is used to estimate the pose from the predicted correspondences.
Finally, A thorough evaluation on LineMod, LineMod-Occlusion, T-Less and Homebrewed DB datasets show our leading performance on the task of weakly-supervised object pose estimation. 

{\small
\bibliographystyle{ieee_fullname}
\bibliography{literature}
}

\end{document}